%% file: main.tex
\documentclass[runningheads]{llncs}
\usepackage[T1]{fontenc}
\usepackage{graphicx,verbatim}
\usepackage{xspace}

\makeatletter
\DeclareRobustCommand\onedot{\futurelet\@let@token\@onedot}
\def\@onedot{\ifx\@let@token.\else.\null\fi\xspace}

\def\eg{\emph{e.g}\onedot} 
\def\ie{\emph{i.e}\onedot}

\makeatother

\usepackage[noadjust]{cite} %

\usepackage{amsmath}

\usepackage{amssymb}
\usepackage{amsfonts}
\usepackage{caption}
\usepackage{subcaption}
\usepackage{hyperref}

\usepackage[capitalize]{cleveref}
\crefname{section}{Sec.}{Secs.}
\Crefname{section}{Section}{Sections}
\Crefname{table}{Table}{Tables}
\crefname{table}{Tab.}{Tabs.}

\usepackage{multirow}
\usepackage{makecell}
\usepackage{utfsym}
\newcommand{\crossmark}{\scalebox{0.75}{\usym{2613}}}

\usepackage{color}

\usepackage[normalem]{ulem}

\begin{document}

\title{U-Mamba2-SSL for Semi-Supervised Tooth and Pulp Segmentation in CBCT}
\titlerunning{U-Mamba2-SSL}
\author{Zhi Qin Tan\inst{1}\orcidID{0000-0002-5521-6808} \and
Xiatian Zhu\inst{2}\orcidID{0000-0002-9284-2955} \and\\
Owen Addison\inst{1}\orcidID{0000-0002-0981-687X} \and
Yunpeng Li\inst{1}\orcidID{0000-0003-4798-541X}}
\authorrunning{Z.Q. Tan et al.}
\institute{Centre for Oral, Clinical \& Translational Sciences, King's College London, United Kingdom\\
\email{\{zhi\_qin.tan,owen.addison,yunpeng.li\}@kcl.ac.uk} \and
Surrey Institute for People-Centred AI, University of Surrey, United Kingdom\\
\email{\{xiatian.zhu,yunpeng.li\}@surrey.ac.uk}}

\maketitle              %
\setcounter{footnote}{0} %

\input{sec/0_abstract}    
\input{sec/1_intro}
\input{sec/2_method}
\input{sec/3_experiments}
\input{sec/4_conclusion}

\bibliographystyle{splncs04}
\bibliography{references}

\clearpage \begin{center} {\Large \textbf{Supplementary Material} \par} \end{center}
\appendix
\input{sec/9_appendix}

\end{document}

%% file: sec/0_abstract.tex
\begin{abstract}

Accurate segmentation of teeth and pulp in Cone-Beam Computed Tomography (CBCT) is vital for clinical applications like treatment planning and diagnosis. However, this process requires extensive expertise and is exceptionally time-consuming, highlighting the critical need for automated algorithms that can effectively utilize unlabeled data.
In this paper, we propose {\bf U-Mamba2-SSL}, a novel semi-supervised learning framework that builds on the U-Mamba2 model and employs a multi-stage training strategy. The framework first pre-trains U-Mamba2 in a self-supervised manner using a disruptive autoencoder. It then leverages unlabeled data through consistency regularization, where we introduce input and feature perturbations to ensure stable model outputs. Finally, a pseudo-labeling strategy is implemented with a reduced loss weighting to minimize the impact of potential errors.
U-Mamba2-SSL achieved an average score of 0.789 and a DSC of 0.917 on the hidden test set, achieving first place in Task 1 of the STSR 2025 challenge.
The code is available at \url{https://github.com/zhiqin1998/UMamba2}.

\keywords{Semi-supervised learning \and U-Mamba2-SSL \and CBCT Imaging \and Tooth and Pulp Segmentation \and STSR 2025 Challenge}

\end{abstract}

%% file: sec/1_intro.tex
\section{Introduction}
Cone-Beam Computed Tomography (CBCT) provides comprehensive 3D information of the oral region and is an important imaging tool in dentistry, as shown by its rapid adoption in dental clinics \cite{clinical_cbct_review}. Precision segmentation of the tooth and pulp structures is vital to various applications such as dental conditions diagnosis, orthodontic procedures, treatment and surgery planning \cite{cbct_use_tyndall2012,cbct_endo}. However, manual segmentation of CBCT scans requires specialized training and is extremely time-consuming due to its high resolution containing a massive number of voxels and the high variability across scans, making it impractical to scale up in practice. This highlights the significance of developing effective semi-supervised approaches with only limited labeled data while leveraging a large amount of unlabeled CBCT scans  \cite{sdtooth,sts2023,ctooth_plus}.

Semi-supervised learning (SSL) incorporates elements from both supervised and unsupervised learning \cite{semisupervised,semisupervised2}, utilizing both labeled and unlabeled data to improve the performance on the supervised task by exploring the latent knowledge from unlabeled data. This alleviates the need for a significant amount of labels, which can require considerable resources to obtain. We focus on three categories of SSL: 
1) Knowledge transfer with pre-training refers to the transfer of knowledge from one task to another via pre-training, where autoencoders \cite{denoising_ae,mae,dae} are trained to reconstruct corrupted input from a large amount of unlabeled data to guide the randomly initialized model weights towards potentially better regions;
2) Consistency regularization training \cite{cct,cr_ssl,cr_vae} based on the smoothness assumption, enforces the model to produce similar output after perturbing the input, internal features, or model weights, pushing the model towards better generalization capability; and
3) Pseudo labeling method \cite{pseudolabel}, one of the most common approaches in SSL due to its simplicity and model-agnostic nature. It is a form of entropy regularization \cite{entropy_reg} with unlabeled data, reducing the overlap of class probability distribution and favoring a low-density class separation.

In this paper, we present U-Mamba2-SSL, a multi-stage semi-supervised learning framework for tooth and pulp segmentation in 3D CBCT images, developed in the scope of the STSR 2025 Task 1 Challenge \cite{stsr2025}.
To exploit the vast amount of unlabeled CBCT data, we first pre-train U-Mamba2 \cite{umamba2} with the disruptive autoencoder on all provided data. Then, the second training stage involves using the labeled data for supervised learning and the unlabeled data for unsupervised learning via consistency regularization techniques in the input and feature spaces. Lastly, the final stage introduces the pseudo labeling method to the training procedure of the previous stage, with a lower loss weight to further optimize the model weights.
The extensive experiments demonstrate the superior performance of our method, outperforming other alternatives and achieving first place with an average score of 0.789 in the STSR 2025 hidden test set.

%% file: sec/2_method.tex
\section{Method}
\begin{figure}[tb]
    \centering
    \includegraphics[width=\linewidth]{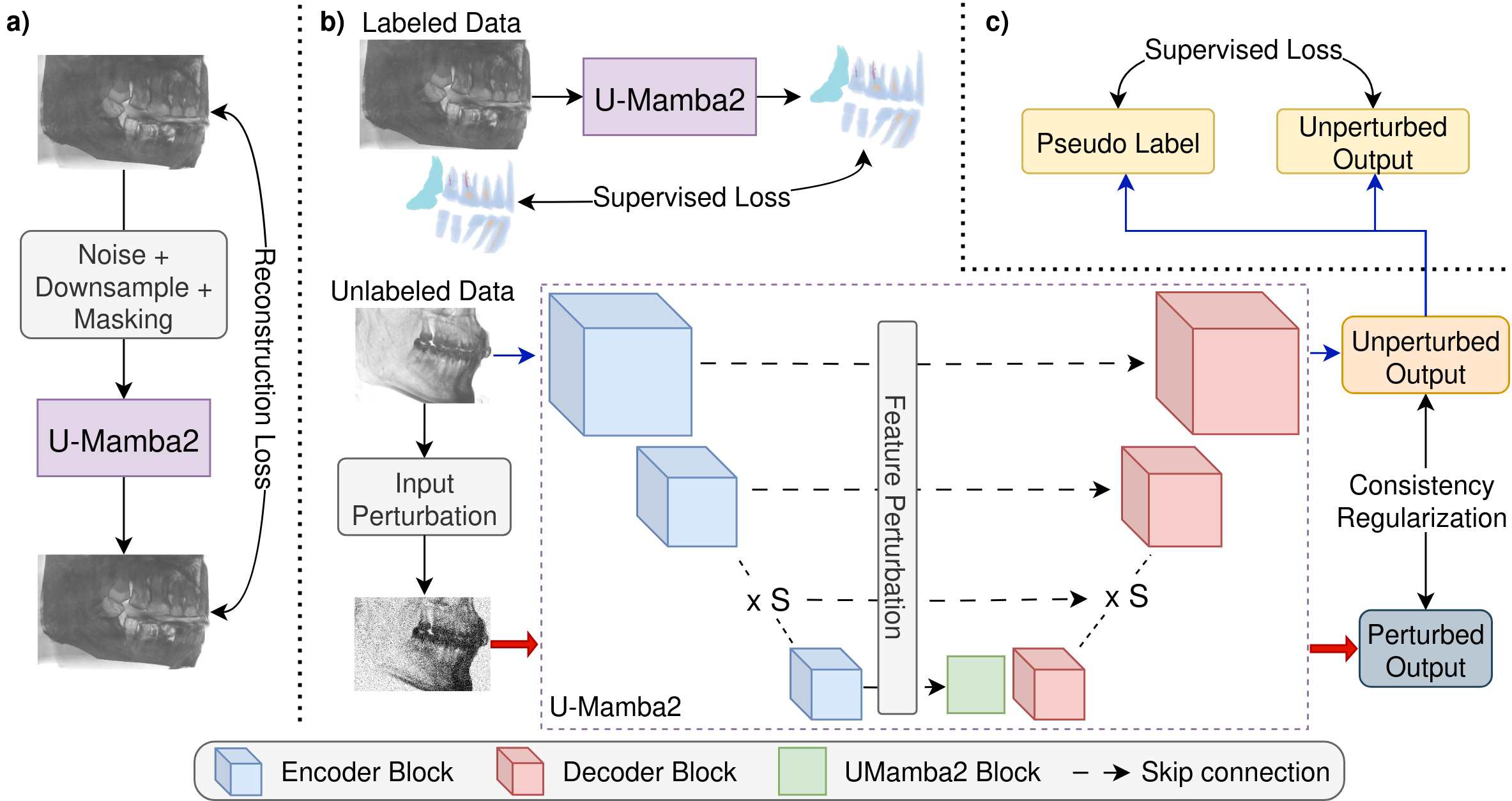}
    \caption{
    Overall diagram of the proposed U-Mamba2-SSL framework. 
    (a) U-Mamba2 is first pre-trained by reconstructing inputs corrupted with noise, downsampled, and masked;
    (b) The second stage involves a combination of supervised loss for the labeled data and consistency regularization between the unperturbed output and the perturbed output of the unlabeled data;
    (c) The final stage introduces pseudo labeling on top of the training objectives in (b). Only pseudo labels with confidence above a certain threshold contribute to the training loss.
    }
    \label{fig:overall_arch}
\end{figure}

\cref{fig:overall_arch} shows the overall process of the U-Mamba2-SSL framework, consisting of three training stages for the U-Mamba2 \cite{umamba2} model. U-Mamba2 integrates Mamba2 \cite{mamba2} state space models into the U-Net architecture at the bottleneck region to enhance its ability to capture long-range dependencies. Mamba2 improves upon Mamba \cite{mamba} by enforcing stronger constraints on the hidden space structure, leading to higher efficiency without compromising its performance compared to transformer-based alternatives. 
We present the details of the three training stages: pre-training, consistency regularization training, and pseudo labeling, in the following subsections.
Note that the final checkpoint of each training stage is used to initialize the model of the subsequent stage.

\paragraph{\textbf{Problem Formulation.}}
Let $\mathcal{D}_l=\{(x_{1}^l, y_{1}), ..., (x_{n}^l, y_{n})\}$ represent the $n$ labeled samples and  $\mathcal{D}_{u}=\{x_1^u, ..., x_m^u\}$ represent the $m$ unlabeled samples, where $x_i^l \in \mathbb{R}^{H \times W \times D}$ is the $i$-th labeled input image, $y_i \in \mathbb{R}^{C \times H \times W \times D}$ is its corresponding voxel-level label, and  $x_i^u$ is the $i$-th unlabeled input image. Here, $C$ is the number of classes while $H, W, D$ are the spatial dimensions. Our goal is to exploit the larger number of unlabeled data (\ie $m \gg n$) to train a 3D segmentation model.

\subsection{First Stage: Pre-training with Disruptive Autoencoder}
\label{subsec:pretrain}
In the medical image domain, data scarcity due to various factors such as complex ethical regulations for accessing and releasing datasets publicly, presents challenges to model pre-training. Therefore, unlike in computer vision tasks of natural images, models for medical image applications are often trained from scratch with random initialization of model weights. However, recent works \cite{dae,Tang2021SelfSupervisedPO} have shown that pre-training deep learning models for medical image tasks can lead to better models that can extract meaningful feature representations to enhance the performance of downstream segmentation tasks, particularly when there is limited labeled data to train from scratch effectively. 

In the first stage of our proposed SSL framework, we utilize all training data (\ie $\mathcal{D}_l \cup \mathcal{D}_u$) to pre-train U-Mamba2 via the disruptive autoencoder (DAE) \cite{dae} method. The DAE method combines three low-level reconstruction tasks for pre-training, namely denoising, super-resolution, and recovering masked information. 

Denoising refers to the task of restoring the original input from its noisy version, obtained by introducing random additive Gaussian noise to the original input. The model must learn to restore all local details in images, such as edges and textures, to output a good denoised image. Besides that, super-resolution is the task of increasing the resolution of a low-resolution image, created artificially by downsampling the original input with linear interpolation. To obtain a good upsampled image, the model must be able to recover the fine details of the image with both local and global information. Lastly, we apply masking to random cubical regions in the input image, setting the voxel values to zero. As most of the information in medical images is not global but is in the finer local details, we use a small cube size relative to the spatial dimensions of the input to prevent discarding too much local information. The model is directed to recover the masked regions, leading to the ability to extract meaningful global context. 
After applying the three input disruptions, U-Mamba2 learns to reconstruct the original image from the corrupted input with an L1 loss function. 

\subsection{Second Stage: Consistency Regularization Training}
We exploit the smoothness assumption and employ consistency regularization training in the second training stage, enforcing the invariance of predictions on the model. In this training stage, we use a combination of supervised loss and unsupervised loss to learn the model parameters. For a labeled training data, $x_{i}^l$, and its voxel-level class label, $y_{i}$, the model is trained in a supervised fashion based on the combination of Dice loss and cross-entropy loss, $\mathcal{L}_{S}$. For an unlabeled training data, $x_{i}^u$, it is first passed through the model to obtain an unperturbed output, $\hat{y}_{i}^u$. Then, we introduce input and feature perturbations \cite{cct} to $x_{i}^u$ and obtain the perturbed output, $\tilde{y}_i^u$, by passing the perturbed input through the model. The semi-supervised consistency regularization loss, $\mathcal{L}_{CR}$, is computed as the $\mathcal{L}_1$ loss between $\hat{y}_{i}^u$ and $\tilde{y}_i^u$. We describe the perturbation details in the following paragraphs.

\subsubsection{Input Perturbations.}
We apply strong data augmentation to the unlabeled data to obtain a perturbed input. It is crucial not to apply spatial (\eg mirroring or rotation) augmentations, as in the context of segmentation, these transformations are non-local and violate the smoothness assumption. Specifically, in this stage, we apply median filter, Gaussian blur, Gaussian noise, random brightness, random contrast, low-resolution simulation, and image sharpening filter.

\subsubsection{Feature Perturbations.}
The perturbed inputs are passed through the encoder blocks in U-Mamba2 to obtain multi-scale 3D feature maps. Before the encoder feature maps are connected to the decoder blocks via skip connections, we apply random perturbations in the feature space to encourage the model to learn more robust and generalizable feature representations. The feature perturbations consist of dropping activations or injecting noise in the encoder feature maps:
\begin{itemize}
    \item Random Spatial Dropout \cite{spatial_dropout}: We apply random channel-wise dropout with a probability of $0.5$. In contrast to i.i.d. dropout, this promotes channel-wise independence in the encoder feature maps.
    \item Random Activation Dropout \cite{dropout}: Activations with high values are randomly dropped to enforce the model to focus on inactive regions in the feature map. We randomly sample a threshold, $\gamma_{drop} \sim \mathcal{U}(0.7, 0.9)$, then set all activations above the $\gamma_{drop}$ percentile to zero. As a result, the top $10\%-30\%$ highly activated regions in the feature map are dropped.
    \item Noise Injection: A noise tensor with the same shape as the feature map is first sampled from a uniform distribution, $N \sim \mathcal{U}(-0.3, 0.3)$. As the activations in the feature maps vary, we ensure that the noise tensor is proportional to the feature map by first multiplying the noise tensor with the feature map before adding it as $Z + (Z \odot N)$, where $Z \in \mathbb{R}^{F \times H \times W \times D}$ is the feature map, $\odot$ is element-wise multiplication, and $F$ is the number of channels.
\end{itemize}

\subsubsection{Semi-Supervised Learning Schedule.}
In practice, we utilize both labeled and unlabeled data during each training epoch. The overall loss signal from both labeled and unlabeled data is computed as 
\begin{equation}
    \mathcal{L} = \mathcal{L}_{S} + \omega_{CR}\mathcal{L}_{CR} \,\,,
    \label{eq:second_stage_loss}
\end{equation}
where $\omega_{CR}$ is the unsupervised loss weight function. $\omega_{CR}$ ramps up exponentially \cite{temp_ensemble} from zero to a fixed weight, $W_{CR}$, at the $0.2T_{ep}$ epoch where $T_{ep}$ is the total number of training epochs. Additionally, we linearly increase the proportion of unlabeled data in each epoch from $10\%$ to $50\%$ at the $0.4T_{ep}$ epoch, allowing the model to focus on learning the main segmentation task in the early phase.

\subsection{Third Stage: Pseudo Labeling}
After the second training stage, we obtain a good U-Mamba2 segmentation model that can maintain local smoothness around its predictions. We capitalize on this feature by further training the model with the pseudo labeling \cite{pseudolabel} strategy. Specifically, the model's predictions on unlabeled data are considered pseudo labels and used for model training in a supervised manner. For the predicted class of each voxel, if the class confidence is above a given confidence threshold, $\lambda_{conf}$, then we use the predicted class as ground truth; otherwise, the voxel is set to the background class and is ignored in the loss calculation. 

In this stage, the loss function from \cref{eq:second_stage_loss} becomes:
\begin{equation}
    \label{eq:third_stage_loss}
    \mathcal{L} = \mathcal{L}_{S} + \omega_{CR}\mathcal{L}_{CR} + W_{PL}\mathcal{L}_{PL} \,\,,
\end{equation}
where $\mathcal{L}_{PL}$ is the supervised loss computed with the pseudo labels and ignores the background class, and $W_{PL}$ is the loss weight for $\mathcal{L}_{PL}$ to balance the loss terms.
Similar to the second stage, we linearly increase the proportion of unlabeled data in each training epoch from $30\%$ to $50\%$ at the $0.2T_{ep}$ epoch.

%% file: sec/3_experiments.tex
\section{Experiments}
\subsection{Dataset and Evaluation Metrics}
The evaluation metrics include Dice Similarity Coefficient (DSC), Normalized Surface Distance (NSD), Mean Intersection over Union (mIoU), and Identification Accuracy (IA) to evaluate the segmentation region overlap and boundary distance. In addition, the algorithm runtime and memory consumption are also evaluated and ranked. 

\subsection{Implementation Details}
\paragraph{\textbf{Preprocessing.}}
We resize all inputs to the median voxel spacing of all training data, $(0.3,0.25,0.25)$, resulting in a median input size of $(337,640,640)$. Then, we clip the input data to the 0.5th and 99.5th percentiles, followed by data normalization based on the mean and standard deviation of the voxel values.

\paragraph{\textbf{Environment settings.}}
The development environments and requirements are presented in Table~\ref{table:env}.

\begin{table}[tb]
    \caption{Development environments and requirements.}
    \label{table:env}
    \centering\setlength{\tabcolsep}{4pt}
    \resizebox{0.66\linewidth}{!}{
    \begin{tabular}{ll}
        \hline
        System & Ubuntu 24.04 \\
        \hline
        CPU & Intel(R) Core(TM) Ultra 9 285K \\
        \hline
        RAM & 2 $\times$ 32GB; 6400 MHz\\
        \hline
        GPU & NVIDIA RTX 5090 32 GB \\
        \hline
        CUDA version & 12.9 \\ 
        \hline
        Programming language & Python 3.11 \\ 
        \hline
        Deep learning framework & PyTorch 2.7.1, nnU-Net 2.6.2 \\
        \hline
    \end{tabular}
    }
\end{table}

\paragraph{\textbf{Training protocols.}}
We implement U-Mamba2-SSL with the nnU-Net \cite{nnunet} framework, using a patch-size training and sliding window inference strategy. During training, we randomly apply rotation, scaling, Gaussian noise, Gaussian blur, brightness and contrast transform, low resolution simulation, and mirroring as data augmentation. The input patch is randomly cropped by ensuring that at least 33\% of the samples have a foreground label. $W_{CR}$, $W_{PL}$, and $\lambda_{conf}$ are set to $50$, $0.1$, and $0.75$, respectively.
All models have 7 encoder-decoder stages and follow the model configuration in \Cref{table:training}.
The provided 30 labeled training data samples are split into 20 training and 10 internal validation splits, where the internal validation split is used to monitor training progress and offline evaluation. We select the checkpoint with the highest DSC on our internal validation set and report the performance metrics on the hidden validation set.

\begin{table}[tb]
    \caption{Training configuration.}
    \label{table:training}
    \centering\setlength{\tabcolsep}{4pt}
    \resizebox{0.66\textwidth}{!}{
    \begin{tabular}{ll} 
        \hline
        Pre-trained Model & See \Cref{subsec:pretrain} \\
        \hline
        Batch size & 2 \\
        \hline 
        Patch size & $128 \times 256 \times 256$  \\ 
        \hline
        Total epochs & 500 \\
        \hline
        Optimizer & SGD with $0.99$ momentum  \\
        \hline
        Initial learning rate  & 0.01 \\ 
        \hline
        Lr decay schedule & Polynomial LR decay \\
        \hline
        Training time & 13 hours \\ 
        \hline 
        Loss function & See \Cref{eq:second_stage_loss,eq:third_stage_loss} \\
        \hline
        Number of model parameters & 156M \\ 
        \hline
        Number of flops & 6.22T\\ 
        \hline
    \end{tabular}
    }
\end{table}

\section{Results and Discussion}
\subsection{Quantitative Results}
\Cref{tab:results} presents the results of our proposed method compared with two baselines, nnU-Net and U-Mamba2. We observe that all methods achieved high DSC, NSD, and mIoU metrics, which measure overall image-level performance. However, U-Mamba2-SSL outperforms others significantly in IA, which calculates the average percentage of classes with IoU $> 0.5$ across all images. Notably, pre-training leads to the largest leap in IA, from 0.464 to 0.731, while incorporating consistency regularization and pseudo labeling further increases IA to 0.738.

\begin{table}[tb]
    \caption{Evaluation results on the validation set. CR is consistency regularization, while PL refers to pseudo label. Higher is better for all metrics.
    }\label{tab:results}
    \centering\setlength{\tabcolsep}{4pt}
    \resizebox{0.85\linewidth}{!}{
    \begin{tabular}{l|ccc|cccc|c}
        \hline
        Methods & Pre-train & CR & PL &   DSC   &   NSD   &   mIoU    &   IA  & Average\\ 
        \hline
        nnU-Net \cite{nnunet} & - & - & - & 0.963 & 0.997 & 0.928 & 0.286 & 0.794 \\
        U-Mamba2 \cite{umamba2} & - & - & - & 0.965 & 0.998 & 0.930 & 0.464 & 0.839 \\
        \hline
        \multirow{3}{*}{U-Mamba2-SSL} & \checkmark & \crossmark & \crossmark & 0.967 & 0.998 & 0.937 & 0.731 & 0.908 \\
        & \checkmark & \checkmark & \crossmark & 0.967 & 0.999 & 0.935 & 0.736 & \textbf{0.910} \\
        & \checkmark & \checkmark & \checkmark & 0.967 & 0.999 & 0.935 & 0.738 & \textbf{0.910} \\
        \hline
    \end{tabular}
    }
\end{table}

\subsection{Qualitative Results}
\begin{figure}[tb]
    \begin{subfigure}{0.49\textwidth}
        \centering 
        \includegraphics[width=0.54\linewidth,trim={7cm 1cm 0 3cm},clip]{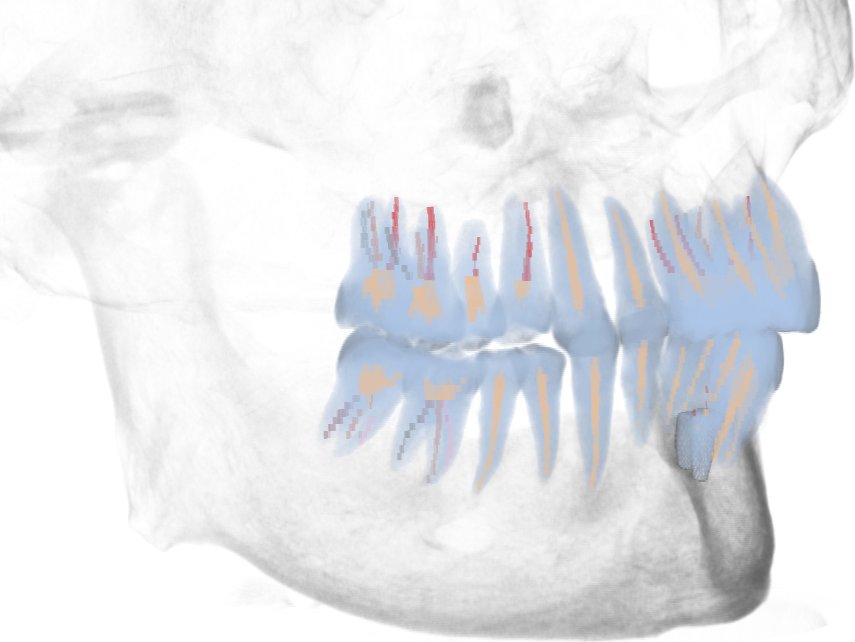}
        \includegraphics[width=0.4\linewidth]{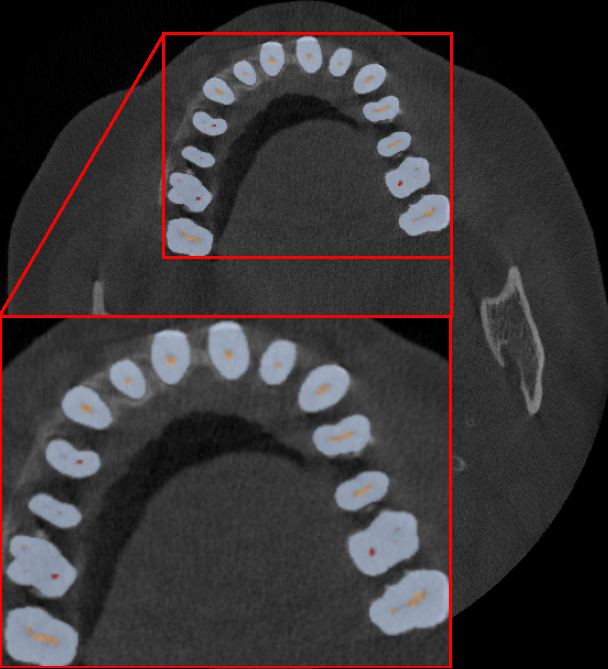}
    \end{subfigure} \hfill
    \begin{subfigure}{0.49\textwidth}
        \centering 
        \includegraphics[width=0.54\linewidth,trim={7cm 1cm 0 3cm},clip]{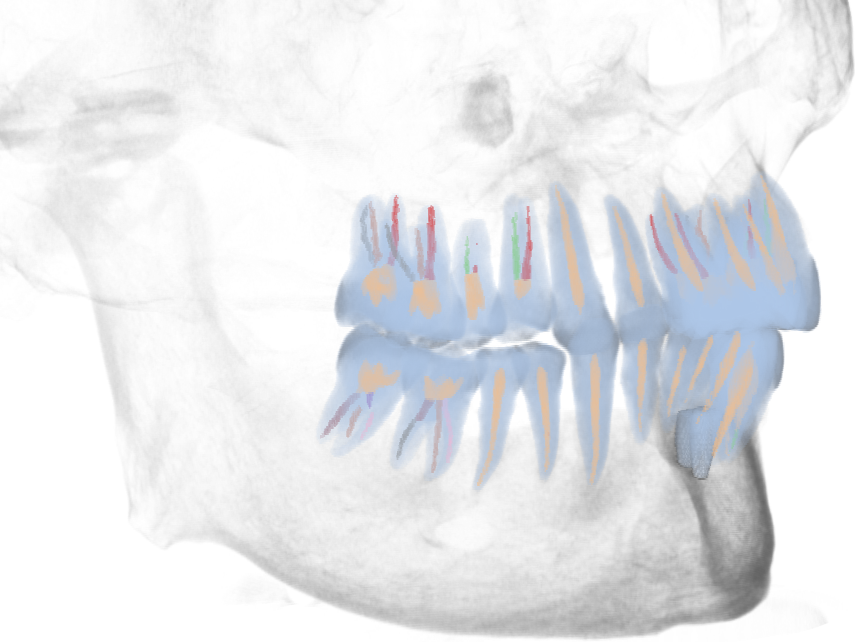}
        \includegraphics[width=0.4\linewidth]{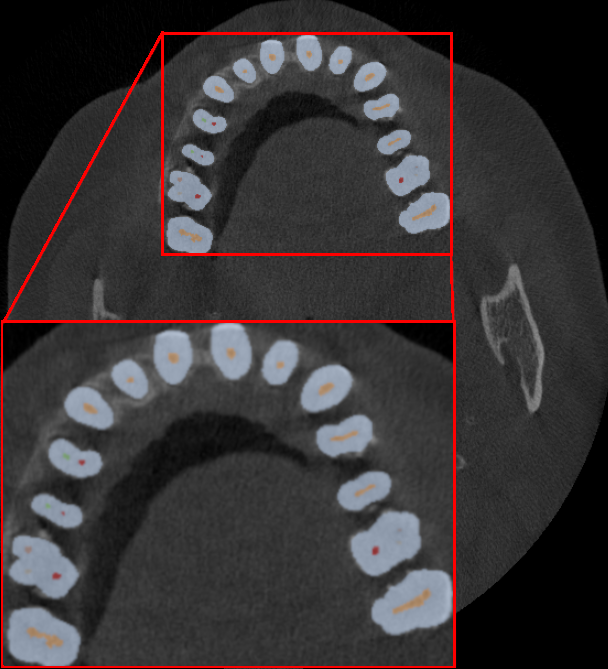}
    \end{subfigure}
    
    \begin{subfigure}{0.49\textwidth}
        \centering 
        \includegraphics[width=0.54\linewidth,trim={0 0 0 2cm},clip]{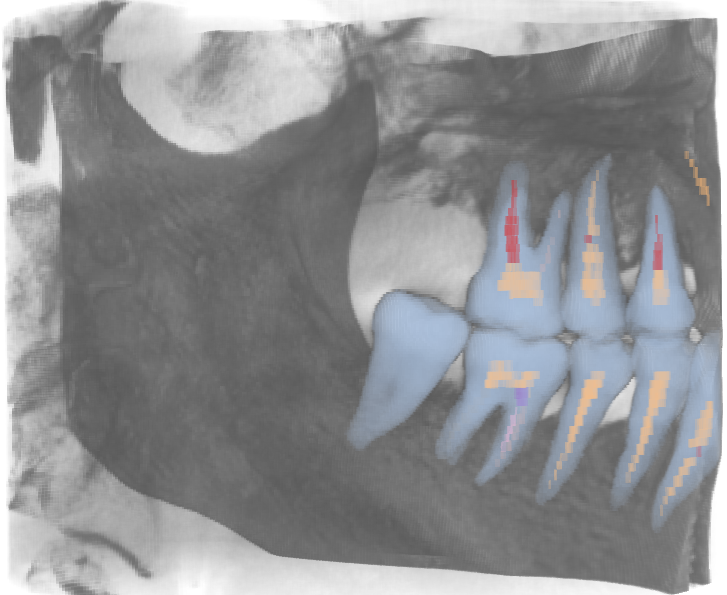}
        \includegraphics[width=0.4\linewidth]{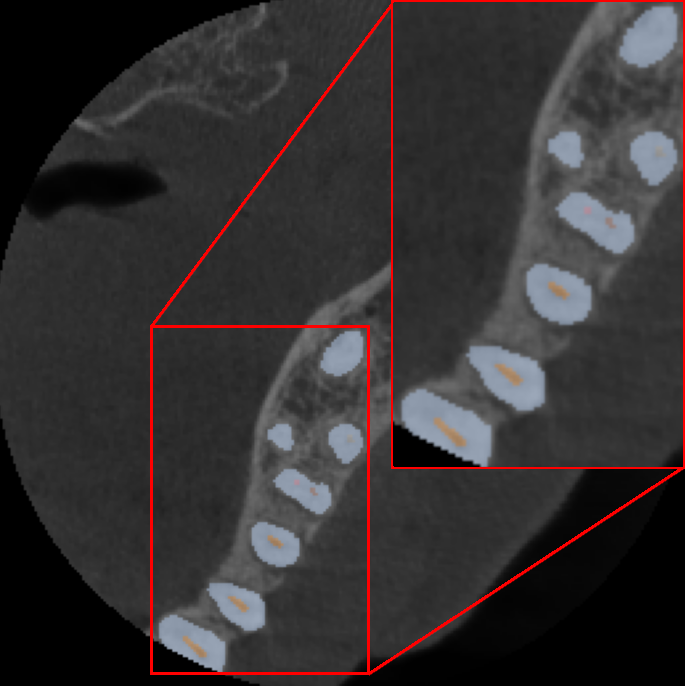}
        \caption{Ground Truth}
    \end{subfigure} \hfill
    \begin{subfigure}{0.49\textwidth}
        \centering 
        \includegraphics[width=0.54\linewidth,trim={0 0 0 2cm},clip]{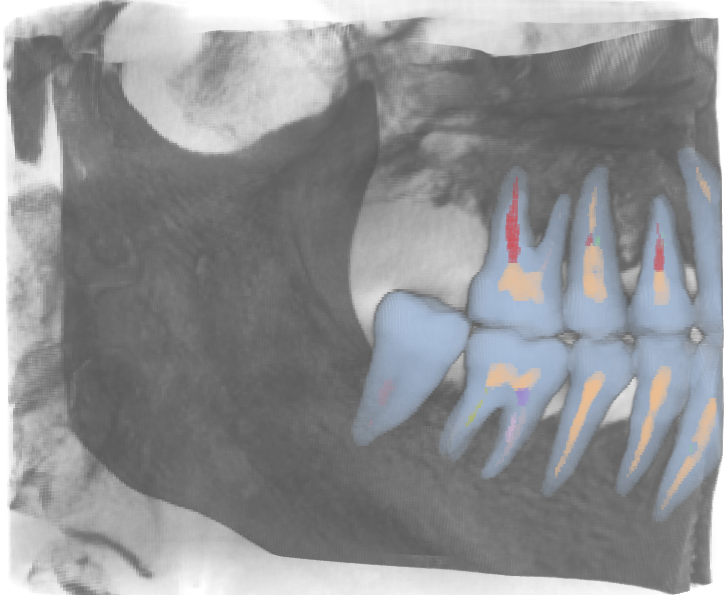}
        \includegraphics[width=0.4\linewidth]{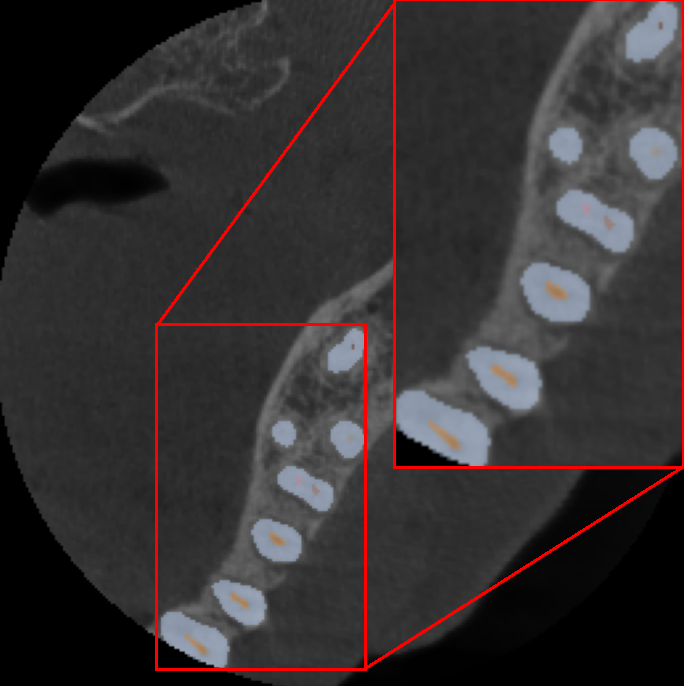}
        \caption{Prediction}
    \end{subfigure}
    \caption{Qualitative results of U-Mamba2-SSL on the internal validation set. The 3D render and a representative 2D slice are shown for: (Top) the best scoring case and (Bottom) the worst scoring case.}
    \label{fig:qualitative}
\end{figure}

\cref{fig:qualitative} shows the qualitative comparison between the ground truth and our model's predictions of the scans with the highest and lowest DSC in our internal validation set, in the top and bottom rows, respectively. Generally, we observe that our method can accurately differentiate between the tooth and different classes of pulp and root canal. The failure cases of our method typically stem from the inability to precisely predict the thickness and the length or extent of the pulp. Moreover, our model also struggles with limited field of view (LFOV) CBCTs where it predicts more false positives around the image edges.

\subsection{Final Challenge Submission}
We scale up our training procedure by training on all available data for 1000 epochs and increasing the input patch size to 160x256x256. For inference, we use a sliding window inference with a tile size of 0.9, and enable mirroring in the anterior/posterior and left/right axes during test-time augmentation (See \Cref{appx:speed} for the speed optimization).
Our method achieved a 0.969 DSC, 0.998 NSD, 0.940 mIoU, and 0.806 IA on the validation set, while obtaining a DSC, NSD, mIoU, and IA of 0.917, 0.882, 0.948, and 0.577, respectively, on the final hidden test set, securing first place in Task 1 of the STSR 2025 challenge.

\subsection{Limitation and Future Work}
Our work, while successful, is not without limitations. Firstly, the dataset consists of full and LFOV CBCTs, which present different information and image properties. Future work should design data processing and augmentation techniques that are specific to the different types of CBCTs to capitalize on their differences. Next, typically, only a small region of interest (ROI) in the CBCT image contains the foreground classes. 
Future research can explore this aspect to avoid spending computational resources and time on volumes with only background classes and allow the model to focus on the ROI.

%% file: sec/4_conclusion.tex
\section{Conclusion}

We presented U-Mamba2-SSL, a novel multi-stage semi-supervised learning framework for tooth and pulp segmentation in CBCT scans, in the scope of the STSR 2025 challenge. The framework consists of first pre-training U-Mamba2 with the disruptive autoencoder, utilizing unlabeled data for consistency regularization, and a pseudo labeling strategy in the final stage.
Our results demonstrate that the proposed framework can substantially enhance model performance, achieving first place with an average score of 0.789 on the hidden test set in Task 1 of the STSR 2025 challenge.

\subsubsection{Acknowledgements} We thank all the data owners for making the medical images publicly available and Codabench~\cite{codabench} for hosting the challenge platform.

\subsubsection{Disclosure of Interests.} The authors have no competing interests to declare that are relevant to the content of this article.

%% file: sec/9_appendix.tex
\section{Optimizing Speed in Sliding Window Inference}
\label{appx:speed}

\begin{figure}[htb]
    \begin{subfigure}{0.52\textwidth}
        \centering 
        \includegraphics[width=\linewidth]{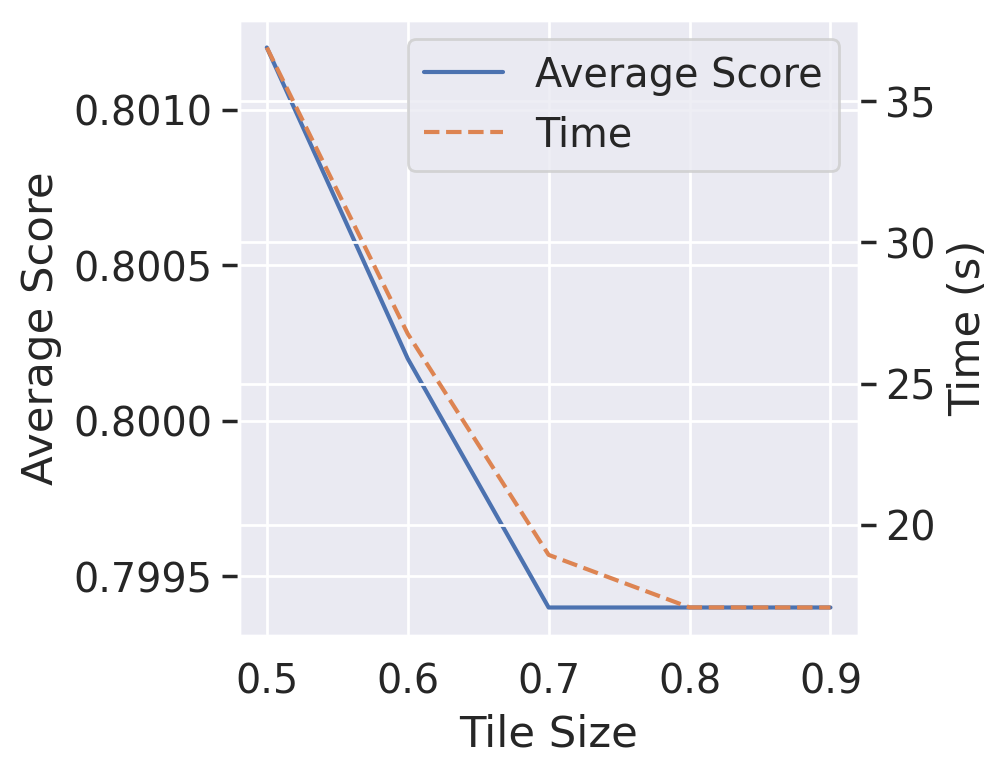}
    \end{subfigure} \hfill
    \begin{subfigure}{0.47\textwidth}
        \centering 
        \includegraphics[width=\linewidth]{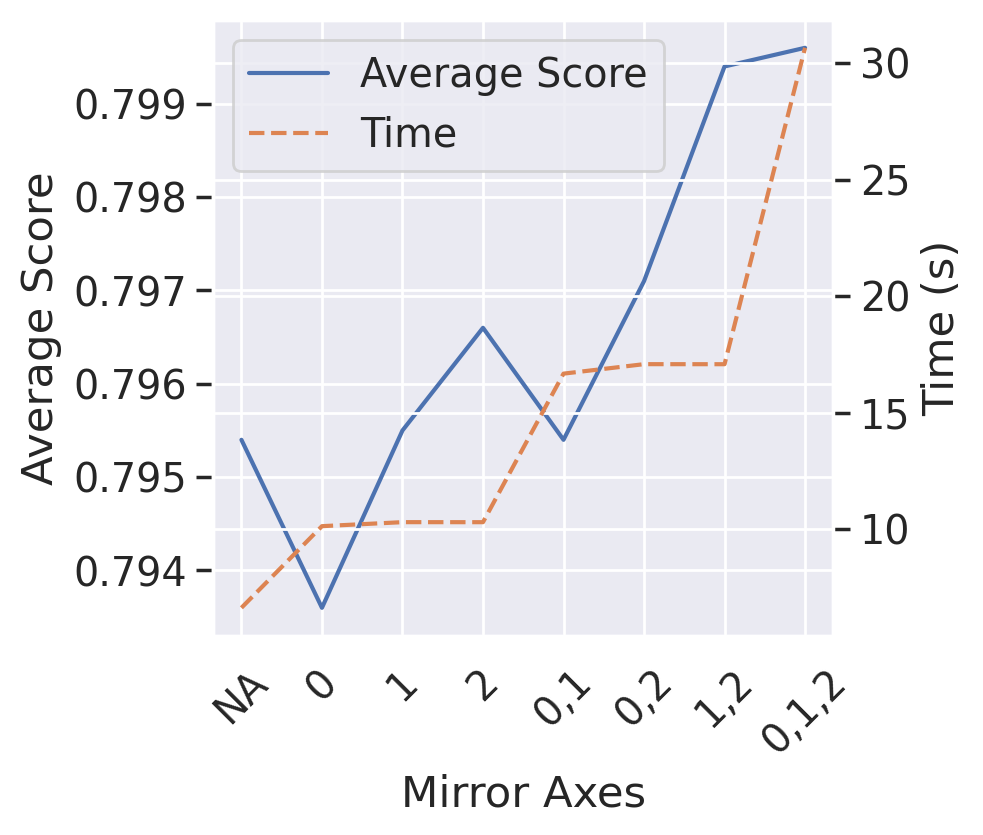}
    \end{subfigure}
    \caption{(Left): Effect of the tile size on the metrics with `1,2' mirror axes in TTA. (Right): Effect of various mirror axes combinations in TTA on the metrics when tile size is set to 0.9. Axis definition: `0' is superior/inferior, `1' is anterior/posterior, and `2' is left/right.}
    \label{fig:speed_tradeoff}
\end{figure}

Since inference time is also an evaluation metric in the STSR 2025 challenge, we explore optimizing the parameters of the sliding window inference technique to improve inference speed without significantly deteriorating model performance (\ie the average score of DSC, NSD, mIoU, and IA). \cref{fig:speed_tradeoff} illustrates the tradeoff between the model performance and inference time for different tile sizes and mirror axes combinations in test-time augmentation (TTA). As most of the voxels in the CBCT image belong to the background class, setting the tile size to 0.9 substantially reduces the inference time by 53\% with a negligible drop of only 0.002 average score. Furthermore, \cref{fig:speed_tradeoff}  demonstrates that although mirroring in all axes leads to the best performance, it comes with the downside of long inference time.
The optimal mirror axes combination is `1,2', offering a good average score with an inference time of only 17.08 seconds. 

